\documentclass{article}
\usepackage{amssymb,amsmath,amsthm}
\usepackage{mathrsfs, bbm, tikz-cd}
\usepackage{bm}
\usepackage{float}
\usepackage{mathtools}

\newcommand{\cC}{\mathcal{C}}
\newcommand{\cE}{\mathcal{E}}

\newcommand{\cL}{\mathcal{L}}

\newcommand{\cO}{\mathcal{O}}

\newcommand{\bbR}{\mathbb{R}}

\providecommand{\norm}[1]{\ensuremath{\left\lVert#1\right\rVert}}

\usepackage{microtype}
\usepackage{graphicx}
\usepackage{subfigure}
\usepackage{booktabs} % for professional tables
\usepackage{float}
\usepackage{hyperref}

\usepackage[accepted]{icml2021}

\icmltitlerunning{}

\begin{document}

\twocolumn[
\icmltitle{Dynamic Customer Embeddings for Financial Service Applications}

% It is OKAY to include author information, even for blind
% submissions: the style file will automatically remove it for you
% unless you've provided the [accepted] option to the icml2021
% package.

% List of affiliations: The first argument should be a (short)
% identifier you will use later to specify author affiliations
% Academic affiliations should list Department, University, City, Region, Country
% Industry affiliations should list Company, City, Region, Country

% You can specify symbols, otherwise they are numbered in order.
% Ideally, you should not use this facility. Affiliations will be numbered
% in order of appearance and this is the preferred way.
\icmlsetsymbol{equal}{*}

\begin{icmlauthorlist}
\icmlauthor{Nima Chitsazan}{equal,c1}
\icmlauthor{Samuel Sharpe}{equal,c1}
\icmlauthor{Dwipam Katariya}{equal,c1}
\icmlauthor{Qianyu Cheng}{equal,c1}
\icmlauthor{Karthik Rajasethupathy}{c1}

\end{icmlauthorlist}

\icmlaffiliation{c1}{Capital One, McLean, VA, USA}

\icmlcorrespondingauthor{Samuel Sharpe}{samuel.sharpe@capitalone.com}
\icmlcorrespondingauthor{Nima Chitsazan}{nima.chitsazan@capitalone.com}
% You may provide any keywords that you
% find helpful for describing your paper; these are used to populate
% the "keywords" metadata in the PDF but will not be shown in the document
\icmlkeywords{Machine Learning, ICML}

\vskip 0.3in
]

% this must go after the closing bracket ] following \twocolumn[ ...

% This command actually creates the footnote in the first column
% listing the affiliations and the copyright notice.
% The command takes one argument, which is text to display at the start of the footnote.
% The \icmlEqualContribution command is standard text for equal contribution.
% Remove it (just {}) if you do not need this facility.

%\printAffiliationsAndNotice{}  % leave blank if no need to mention equal contribution
\printAffiliationsAndNotice{\icmlEqualContribution} % otherwise use the standard text.

\begin{abstract}

As financial services (FS) companies have experienced drastic technology driven changes, the availability of new data streams provides the opportunity for more comprehensive customer understanding. We propose Dynamic Customer Embeddings (DCE), a framework that leverages customers' digital activity and a wide range of financial context to learn dense representations of customers in the FS industry. Our method examines customer actions and pageviews within a mobile or web digital session, the sequencing of the sessions themselves, and snapshots of common financial features across our organization at the time of login. We test our customer embeddings using real world data in three prediction problems: 1) the intent of a customer in their next digital session, 2) the probability of a customer calling the call centers after a session, and 3) the probability of a digital session to be fraudulent. DCE showed performance lift in all three downstream problems.

\end{abstract}

\section{Introduction}

The digitization and wide-spread adoption of technology in the financial services (FS) industry has led to an abundance of financial products for customers. Credit cards, rewards, loans, banking, investing, and budgeting are just some of the services individuals have at their fingertips. This transformation has prompted an explosion of new data streams for FS companies that capture interactions with customers and users' implicit intent in all areas of their financial management environment. In order to practically leverage the temporal dynamics of user behavior for a wide variety of prediction tasks in the banking ecosystem, there is a need for compact and dynamic representations of customer activities.

Well designed representation methods as seen in natural language processing or computer vision may be able to better capture customer behavior rather than relying mostly on feature engineering and aggregated account data. Recent work in the FS industry has embraced granularity of available transaction data with dynamic, task specific approaches for fraud \cite{branco2020interleaved} and credit risk \cite{clements2020sequential}. Other approaches utilize transaction data to generate static general purpose customer embeddings for downstream tasks through graph techniques \cite{khazane2019deeptrax} and autoencoders \cite{baldassini2018client2vec}. Currently, representation learning methods for modeling customer activity in the FS industry either only use transaction and account data, are designed for single tasks, or are static and cannot effectively model changes in user activity. 

Therefore, we propose a framework we refer to as Dynamic Customer Embeddings (DCE) that condenses customer online activity and fuses with just in time financial context to create dynamic customer representations for various downstream tasks. We see that combining online activity with this account context improves performance over models that only utilize customers' financial activity.

\section{Related Work}

Dynamic representation learning for user activities has been addressed in other industries to predict social media user engagement \cite{tang2020knowing, liu2019characterizing}, assess customer value or churn \cite{yang2018know, chamberlain2017customer}, predict conversion \cite{zhou2019understanding}, and forecast student actions in massive open online courses \cite{pandey2020learning}. A majority of the work in customer representation learning, however, has been focused on recommendation systems \cite{wang2020survey}. Most recently, recommender systems that are session based or adapt to new user behavior have gained popularity due to benefits over methods that only learn static user preferences.

These dynamic recommender systems have benefited from the emergence of recurrent neural networks (RNN). \citet{RRN} utilized Long Short Term Memory (LSTM) networks to estimate the movie ratings, outperforming collaborative filtering (CF) methods with hand-engineered temporal corrections \cite{koren2009collaborative}. \citet{dai2016deep} parameterized multidimensional point processes with RNNs in DeepCoevolve to surpass performance of CF and traditional point process recommendation techniques on a variety of recommendation tasks. 

Alterations to RNN architectures have also been proposed to more effectively model time and context that accompanies interactions between users and items. TimeLSTM introduced time gates to better capture long and short term interests of users \cite{zhu2017next}. LatentCross proposed an element-wise multiplication of context and hidden states to effectively incorporate context features in recurrent recommendation systems \cite{beutel2018latent}.

Recently, \citet{kumar2019predicting} developed JODIE which included two important and unique concepts. First, they proposed mutually recursive RNNs which produce dynamic embeddings for both users and items. Previous RNN approaches such as TimeLSTM and LatentCross only considered users' changing preferences. Second, they included a projection operator whereby RNNs predict the future user embeddings. JODIE is able to use the projected embedding to identify the most similar items by embedding distance, reducing inference time from previous methods which needed to estimate a probability for each item.

Other methods have built off the successful concepts established by JODIE. \citet{kefato2020dynamic} reduced the need for complex batch design by using attention on short term interactions of users and items to project user embeddings. \citet{pandey2021iacn} pivoted to an attention based architecture and included an influence layer that allows a user's neighborhood to increasingly influence their embedding in the presence of inactivity. Most relevant to our work, \citet{hansen2020contextual} applies JODIE's embedding prediction to their music recommendation task. Since music tracks are often consumed together in a session, they posit that sequences of sessions capture the true change in user preferences. Given that sessions can each be unique, they use a word2vec representation \cite{mikolov2013efficient} of tracks to create session embeddings which they use in place of traditional item embeddings. 

In contrast to these applications, FS companies retain a wider range of customer contextual information which is predictive of customer behaviour (see Table~\ref{tab:context} for examples). We propose the DCE framework to segregate the contextual features in multiple groups based on their temporal characteristics and fuse the resulting vectors from each group to form the overall customer embeddings. Our results show that the embeddings generated by this framework outperform the other methods in predicting the next customer action and improve internal decisioning systems. To the best of our knowledge, this is the first study in the FS industry that integrates click-stream data with customers' financial information to generate customer level embeddings for utilization in downstream tasks. 

In the following sections we describe our internal datasets, the problem setup and representation learning framework, and results on internal customer servicing and fraud prediction tasks.

\section{Data}
% We use data internal to our company, a Fortune 500 bank with millions of customers, for our experiments. The data sources range from customers' online activities on web and mobile apps to financial statement and account information. 

We use data internal to Capital One for our experiments. The data sources range from customers' online activities on web and mobile apps to financial statement and account information. 

\subsection{Online/Click-Stream Data}
To capture the online activities we use click-stream data after users login to web or mobile platforms. We collect click-stream data in its most granular elements: pageviews or actions. In this study we refer to the pageviews or actions as click-stream events. Each event is associated with a timestamp. The sequence of click-stream events that starts with login and ends with the logout is referred to as a \textit{digital session}.

\subsection{Financial/Account Context}
Snapshots of contextual financial and account information, available at each session, provide relevant information about the customer's financial state. These snapshots are produced at the time of login before any other actions are taken. Table~\ref{tab:context} lists these features.

\begin{table}[!htbp]
\caption{Financial context feature categories}
\label{tab:context}
\vskip 0.15in
\begin{center}
\begin{tabular}{ccc}
\toprule
Context Feature Type & \# of Features \\
\midrule
Posted Transactions & 10 \\
Transactions Authorization & 47\\ 
Account & 29 \\
Utilization & 20\\
Payments & 12  \\
Rewards & 10 \\
Digital Massaging & 6  \\
Fraud & 2  \\
\bottomrule
\end{tabular}
\end{center}
\vskip -0.1in
\end{table}

% \begin{table*}[!htbp]
% \caption{Session $i$ generated contextual features}
% \label{sample-table}
% \vskip 0.15in
% \begin{center}
% \begin{tabular}{ccc}
% \toprule
% Context Feature Type & \# of Features &  Description \\
% \midrule
% Posted Transactions & 10 &  Stats of recent transactions\\
% Transactions Authorization & 47 & Status of recent recent transactions\\ 
% Account & 29 & Type of customer accounts\\
% Utilization & 20 & Stats of customer utilization's of their accounts \\
% Payments & 12 & Payement related information, e.g. amount of the last payment \\
% Rewards & 10 &  Rewards related information and indicators e.g. total rewards balance\\
% Digital Massaging & 6 & Latest digital messaging with the customer \\
% Fraud & 2 & Time since latest fraud alert \\
% \bottomrule
% \end{tabular}
% \end{center}
% \vskip -0.1in
% \end{table*}

\section{Digital Session Embedding}
Similar to \citet{hansen2020contextual}, we consider the smallest action elements to be more informative when viewed together rather than individually. Therefore, we create an aggregate embedding for each session. 

Formally, we define the set of $m$ click-stream events as $\cE = \{e_j \; | \; j = 1 \dots m\}$. Sessions are composed of a time ordered collection of $K$ events $S=\{e^k_j\;|\; e_j \in \cE, \; \; t^k< t^{k+1},\; k = 1 \dots K\}$.

We use a standard sequence to sequence LSTM autoencoder to learn representative embeddings $s \in \bbR^d$ for the digital session. The output of the LSTM encoder of each session becomes the embedding for that digital session. 

\section{Dynamic Customer Embeddings (DCE)}
In this section, we describe DCE, a self-supervised method to learn online customer activity embeddings. 

\subsection{Setup}
We observe a customer $c\in \cC$ logging into the web or mobile app to start an online session $S_i$ at time $t_i \in \bbR^+$. For each session $S_i$ we produce a corresponding embedding $s_i$ as described in the previous section. At the time of login we are given a snapshot of their financial context $f_i \in \bbR^F$. The set of observations for a user $u$ can be represented by the set of tuples $\cO^u = \{(t_i, s_i, f_i)|\;t_{i-1} < t_i\;i = 1 \dots N_c\}$. 

In order to effectively encode customer behavior and intent, DCE attempts to predict the activity in the next session, given past customer digital activity up to and including the previous session and previous financial activity and other context at the time of login. The features associated with each session are detailed in Table~\ref{tab:sessfeat}.
\begin{table}[!htbp]
\caption{Session $i$ features}
\label{tab:sessfeat}
\vskip 0.15in
\begin{center}
% \begin{small}
% \begin{sc}
\begin{tabular}{cc}
\toprule
Notation & Description \\
\midrule
$s_i$    &  session $i$ embedding \\
$\Delta_i$ & $t_i - t_{i-1}$ \\
$D_i$ & day of the week embedding \\
$W_i$ & week of the month embedding\\
$M_i$ & month of the year embedding\\
$f_i$ & financial context at $t_i$ \\
\bottomrule

\end{tabular}
% \end{sc}
% \end{small}
\end{center}
\vskip -0.1in
\end{table}

\subsection{Model}

\begin{figure*}
\begin{minipage}{.32\linewidth}
\centering
\includegraphics[width=\linewidth]{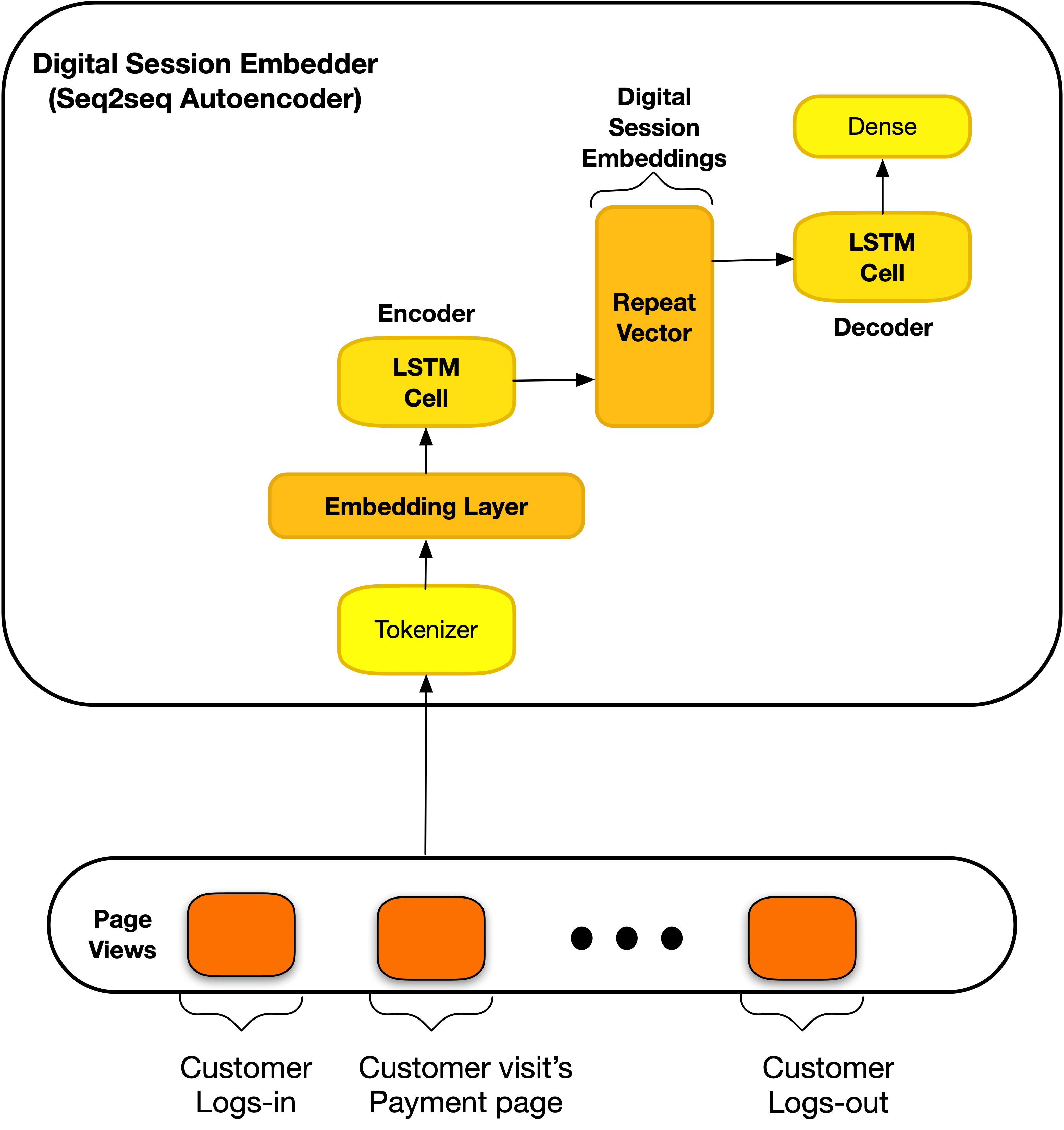}
(a) Session embedding architecture
\end{minipage} \hfill
\begin{minipage}{.60\linewidth}
\centering
\includegraphics[width=\linewidth]{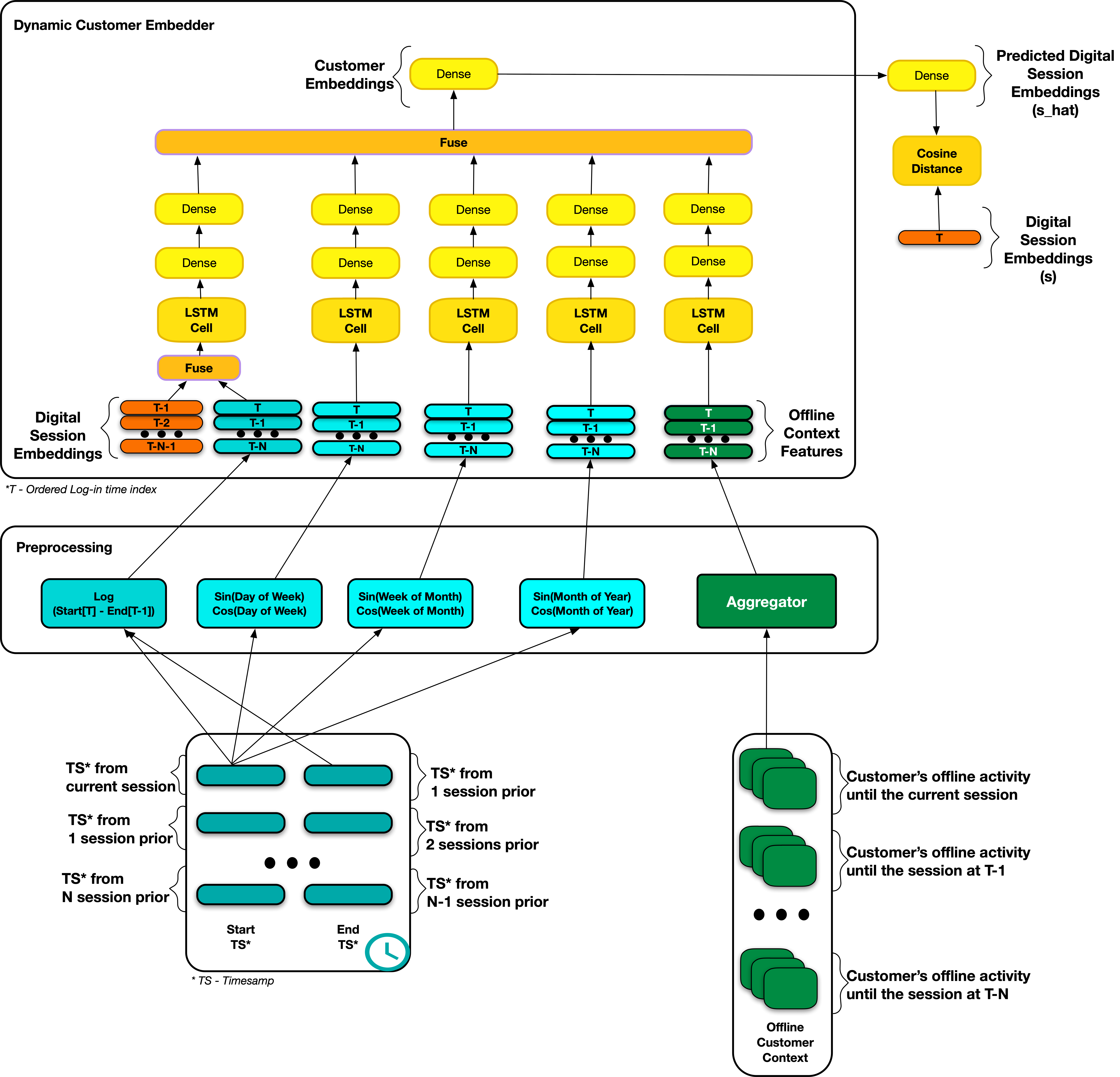}
(b) Customer embedding architecture
\end{minipage}
\caption{The DCE framework combines digital session embeddings of click-stream actions, multiple time representations, and an aggregation of customer context up until the current session.}
\label{fig:arch}
\vskip -0.1in
\end{figure*}

We find that recent recurrent recommendation techniques for short or long term user attention, time projection, and context fusion provide little benefit for our data. Given the complex, high dimensional, and evolving nature of our financial context we split our problem into separate recurrent modules which are fused to form the customer embedding. 
\begin{table}[!htbp]
\caption{LSTM Modules}
\label{tab:lstm}
\vskip 0.15in
\begin{center}
% \begin{small}
% \begin{sc}
\begin{tabular}{ccc}
\toprule
Notation & Input Features \\
\midrule
$LSTM_s$&   $[s_{i-1},\; log(\Delta_i)]$ \\
$LSTM_D$&  $D_i$ \\
$LSTM_W$&  $W_i$ \\
$LSTM_M$&  $M_i$ \\
$LSTM_f$ & $f_i$ \\
\bottomrule

\end{tabular}
% \end{sc}
% \end{small}
\end{center}
\vskip -0.1in
\end{table}
We use five LSTMs detailed in Table~\ref{tab:lstm}, one to update the sequential representation for each separate concept. When a customer logs in to begin session $i$ at $t_i$, we update $LSTM_s$ with the previous session embedding $s_{i-1}$ concatenated with a transformation of the time elapsed since the previous session, $\Delta_i$. We separately update embeddings for the other cyclical and financial information as of $t_i$. 

The updated hidden state $h_{i,k}$, for each module $k$ and observation $i$, is fed through an MLP, concatenated, and passed through a final fully connected (FC) layer representing the customer embedding $c_i$.
\begin{align*}
    o_{i,k} & = MLP(h_{i,k}) \\
    c_i & = FC([o_{i,s}, o_{i,D}, o_{i,W}, o_{i,M}, o_{i,f}])
\end{align*}

Finally, to predict the next session, we simply project the customer embedding into the session embedding space:
\begin{align*}
    \hat{s_i} & = FC(c_i)
\end{align*}

See Figure~\ref{fig:arch} for details on our architecture.  We train DCE in a self-supervised manner by minimizing the cosine distance between the predicted and actual session embedding. We define the loss as
\begin{align*}
   \cL & = \sum_{c \in \cC} \sum_{i=1}^{N_c} 1-\frac{s_i^\top \hat{s_i}}{\norm{s_i}\norm{\hat{s_i}}}.
\end{align*}

\section{Experiments}
For the following experiments we used the digital sessions of a sample of customers in a one year period to train DCE and sessions from the same customers in a subsequent two month period for validation. Finally, an out of sample set of customers were used as the test set. We compare our method to a few naive baselines on the target task of next session embedding prediction:

\begin{itemize}
    \item Previous session embedding: $s_{i-1}$
    \item Average session embedding: $\frac{1}{i-1}\sum_{k=1}^{i-1} s_k$
    \item Exponential moving average session embedding (EMA):
    $$\alpha [ s_{i-1} + (1-\alpha)s_{i-2} \dots (1-\alpha)^{i-2}s_1 ] $$
\end{itemize}
where $\alpha$ is found using gradient descent on the training set. We also include comparisons of four relevant recurrent recommendation architectures in Table~\ref{tab:cosloss}. 

We find that all RNN architectures outperform the baselines and DCE shows an improvement over recent methods. We hypothesize that changes in the financial state of customers is an important element influencing future sessions. Our separate sequential modeling treatment of financial context may be able to capture these dynamics more effectively than previous methods that either concatenate this information in a single LSTM or fuse with the hidden state. 

\begin{table}[h]
\caption{Average cosine distance between predicted and actual session embedding}
\label{tab:cosloss}
\vskip 0.15in
\begin{center}
\begin{small}
% \begin{sc}
\begin{tabular}{cc}
\toprule
Model & Cosine Distance \\
\midrule
Previous session embedding &  0.2107 \\
Average session embedding & 0.1427 \\
Exponential Moving Average & 0.1248 \\
Time LSTM \cite{zhu2017next} & 0.1176 \\
LatentCross \cite{beutel2018latent} & 0.1179 \\
JODIE \cite{kumar2019predicting} & 0.1185 \\
CoSeRNN \cite{hansen2020contextual} & 0.1185 \\
DCE & \textbf{0.1168} \\
\bottomrule
\end{tabular}
% \end{sc}
\end{small}
\end{center}
\vskip -0.1in
\end{table}

\subsection{Customer Intent Prediction}
Predicting customer intents in a digital session is an important step in anticipating user needs and providing a seamless customer experience in digital platforms. In this section we evaluate our method on a customer intent prediction task at the time of login to the bank website or mobile app. 

Table~\ref{tab:intentdist} shows the 16 primary customer intents. Each session can have multiple intents, so we treat this as a multi-label classification problem and evaluate predictions with a macro AUROC score across intent classes. Any session without a clear intent defaults to \textit{Account Summary}.

We fit a classifier on DCE to compare to other internal baselines detailed in Table~\ref{tab:intentauc}. We include the previous intent model which only utilized the contextual features (\textit{Context only}). We also train a vanilla LSTM instead of using the five separate LSTM modules in DCE. DCE outperforms both of these baselines by almost 10 percent. In DCE+C we re-append context with the DCE embeddings resulting in further improvement. The effectiveness of re-concatenated raw context features in DCE+C demonstrates that this information is still not \textit{fully} captured by DCE for specific downstream tasks. This supports previous findings that suggest inclusion of context at multiple points in the pipeline improves performance for recurrent recommendation tasks \cite{smirnova2017contextual}. 

\begin{table}[!htbp]
\caption{Online Intents}
\label{tab:intentdist}
\vskip 0.15in
\begin{center}
\begin{small}
\begin{tabular}{cc}
\toprule
Intent \\
\midrule
Credit Report \\
Deposit  \\
Overdraft Settings \\
Bank Transactions  \\
Account Summary  \\
Transaction Management \\
Statements and Documents  \\
Activate  \\
Redeem with bank  \\
Non Purchase Transaction  \\
Alter Production Terms  \\
Payment  \\
Authorized User  \\
Replace Card  \\
Checks  \\
Account Update  \\
\bottomrule
\end{tabular}
\end{small}
\end{center}
\vskip -0.1in
\end{table}

\begin{table}[h]
\caption{Macro AUROC for predicting customer intents}
\label{tab:intentauc}
\vskip 0.15in
\begin{center}
% \begin{small}
\begin{tabular}{cc}
\toprule
Model & Macro AUROC \\
\midrule
Context only & 0.664 \\
LSTM & 0.689 \\
DCE & 0.758 \\
DCE+C & \textbf{0.785} \\
\bottomrule
\end{tabular}
% \end{sc}
% \end{small}
\end{center}
\vskip -0.1in
\end{table}

\subsection{Call}

A small subset of sessions on our digital platforms lead to customers calling call centers for further assistance. Triaging, mitigating, and reducing these calls plays an important role in providing a better customer experience and reducing the overhead cost of the call centers. Predicting whether a customer is going to call after a digital session is the first step in this process. 

We consider each digital session where a customer places a call in the following six hour period as our target class. Any session that does not fit this criteria receives a no call label.

We trained a classifier to predict these labels for 3 scenarios. In the first two scenarios we only use either the context or session embedding of the last session. In the third scenario, we concatenate the customer embeddings generated by DCE to see the impact of users' historic behaviours on their propensity to call. The results in Table~\ref{tab:call auc} show the importance of the most recent session when predicting calls. The historical trajectory of the customer encoded in DCE helps improve call prediction by 4 percent. 

\begin{table}[!htbp]
\caption{Call Prediction}
\label{tab:call auc}
\vskip 0.15in
\begin{center}
% \begin{small}
\begin{tabular}{cc}
\toprule
Classifier features & AUROC \\
\midrule
Context only & 0.66 \\
$s_i$ & 0.74 \\
$s_i$ + DCE & \textbf{0.77} \\
\bottomrule
\end{tabular}
% \end{sc}
% \end{small}
\end{center}
\vskip -0.1in
\end{table}

\subsection{Fraud}

The embeddings using the DCE were used in the Probabilistic Risk Engine (PRE) for Account Take Over (ATO) fraud detection. ATO fraud occurs when fraudsters attempt to login to a customer account using stolen credentials. The goal of PRE is to evaluate the risk of a session being fraudulent at the point of login and trigger a multi-factor authentication if warranted. Fraud labels are generated by looking at fraud reports from multiple lines of business (LOB). PRE utilizes a random forest to classify sessions as fraudulent or not.  The previous version of PRE used customer characteristics to identify ATO fraud. We hypothesize that fraudsters' behaviour online are significantly more informative than solely customer characteristics. Thus, DCE embeddings are good predictors of the ATO. Table~\ref{tab:fradurecall} shows the results for three versions of the PRE. The first version is a rule based system, the second is PRE without and the third version is the PRE utilizing the customer embeddings. We see that with a fixed login challenge or intervention rate, including DCE results in a 21 percent higher recall than PRE.

\begin{table}[!htbp]
\caption{ATO fraud recall at fixed login challenge rate}
\label{tab:fradurecall}
\vskip 0.1in
\begin{center}
\begin{tabular}{cc}
\toprule
Model &  Recall\\
\midrule
Rule Based  &  26\% \\
PRE & 38\%   \\
PRE+DCE  & \textbf{46}\% \\
\bottomrule
\end{tabular}

\end{center}
\vskip -0.2in
\end{table}

\section{Discussion \& Future Work}

In our work, we have proposed a framework to combine online customer activity data with our widely used internal financial data to improve a variety of customer servicing tasks and fraud prediction. We have shown that recent user embedding techniques may not be as effective in areas of user representation learning where context is evolving, highly dimensional, and contains rich information about the customer from different sources. Though our approach to our specific data and problem outperforms other recurrent recommendation methods on our data, financial context is not fully captured by our framework as illustrated by our intent prediction experiments. 

We emphasize the need for better context fusion and tabular representation learning in the field of recommendation systems especially in the financial service industry. Incorporating additional concepts from contextual forecasting \cite{lim2019temporal,eisenach2020mqtransformer},  multi-modal fusion \cite{sankaran2021multimodal}, tabular self-supervised learning \cite{yoon2020vime,arik2020tabnet}, and multi-task learning will be crucial in combining sequential representations with traditional tabular financial data for diverse downstream tasks. Furthermore, as financial service companies begin to embrace sequential representations of transaction and account activity \cite{branco2020interleaved,clements2020sequential}, joint learning of online and financial activity will lead to even better representations of customers.

\bibliography{main}
\bibliographystyle{icml2021}

%%%%%%%%%%%%%%%%%%%%%%%%%%%%%%%%%%%%%%%%%%%%%%%%%%%%%%%%%%%%%%%%%%%%%%%%%%%%%%%
%%%%%%%%%%%%%%%%%%%%%%%%%%%%%%%%%%%%%%%%%%%%%%%%%%%%%%%%%%%%%%%%%%%%%%%%%%%%%%%

\end{document}